\documentclass{bmvc2k}

\usepackage{multirow}
\usepackage{makecell}
\usepackage{amssymb}
\usepackage{soul}

\def\titlespace{\hspace{0.4cm}}

\soulregister\cite7
\soulregister\ref7
\soulregister\pageref7

\title{\titlespace Wide Residual Networks}

\addauthor{Sergey Zagoruyko}{sergey.zagoruyko@enpc.fr}{1}
\addauthor{Nikos Komodakis}{nikos.komodakis@enpc.fr}{1}

\addinstitution{
 Universit\'e Paris-Est, \'Ecole des Ponts ParisTech\\
 Paris, France
}

\runninghead{Sergey Zagoruyko and Nikos Komodakis}{Wide residual networks}

\def\eg{\emph{e.g}\bmvaOneDot}

\newcommand{\ve}[1]{\mathbf{#1}} 

\begin{document}

\maketitle

\begin{abstract}
  Deep residual networks were shown to be able to scale up to thousands of layers and still have improving performance. However, each fraction of a percent of improved accuracy costs nearly doubling the number of layers, and so training very deep residual networks has a problem of diminishing feature reuse, which makes these networks very slow to train. To tackle these problems, in this paper we conduct a detailed experimental study on the architecture of ResNet blocks, based on which we propose a novel architecture where we decrease depth and increase width of residual networks. We call the resulting network structures wide residual networks (WRNs) and show that these are far superior over their commonly used thin and very deep counterparts. For example, we demonstrate that even a simple 16-layer-deep wide residual network outperforms in accuracy and efficiency all previous deep residual networks, including thousand-layer-deep networks, achieving new state-of-the-art results on CIFAR, SVHN, COCO, and significant improvements on ImageNet. Our code and models are available at \url{https://github.com/szagoruyko/wide-residual-networks}.
\end{abstract}

\section{Introduction}
\label{sec:intro}

Convolutional neural networks have seen a gradual increase of the number of layers in the last few years, starting from AlexNet \cite{AlexNet}, VGG \cite{Simonyan15}, Inception \cite{GoogLeNet} to Residual \cite{he2015deep} networks, corresponding to improvements in many image recognition tasks. The superiority of deep networks has been spotted in several works in the recent years \cite{Bianchini,MontufarPCB14}. However, training deep neural networks has several difficulties, including exploding/vanishing gradients and degradation. Various techniques were suggested to enable training of deeper neural networks, such as well-designed initialization strategies \cite{GlorotAISTATS2010,journals/corr/HeZR015}, better optimizers \cite{SutskeverMartensDahlHinton_icml2013}, skip connections \cite{dsn,AISTATS2012_RaikoVL12}, knowledge transfer \cite{Romero-et-al-TR2014,Chen:ICLR16} and layer-wise training \cite{Schmidhuber:92ncchunker}.

The latest residual networks \cite{he2015deep}  had a large success winning ImageNet and COCO 2015 competition and achieving state-of-the-art in several benchmarks, including object classification on ImageNet and CIFAR, object detection and segmentation on PASCAL VOC and MS COCO. Compared to Inception architectures they show better generalization, meaning the features can be utilized in transfer learning with better efficiency. Also, follow-up work showed that residual links speed up convergence of deep networks \cite{inception-v4}. Recent follow-up work explored the order of activations in residual networks, presenting identity mappings in residual blocks \cite{basicblock2} and improving training of very deep networks. Successful training of very deep networks was also shown to be possible through the use of highway networks  \cite{highway}, which is an architecture  that had been proposed prior to residual networks. The essential difference between residual and highway networks is that in the latter residual links are gated and weights of these gates are learned.

Therefore, up to this point, the  study of residual networks has focused mainly  on the order of activations inside a ResNet block and the depth of residual networks.  In this work we attempt to conduct an experimental study that goes beyond the above points. By doing so, our goal is  to explore a much richer set of  network architectures  of  ResNet blocks and thoroughly examine how several other different aspects besides the order of activations affect performance. As we explain below, such an exploration of architectures has led to   new interesting findings  with great practical importance concerning residual networks.

\textbf{Width vs depth in residual networks}.
The problem of shallow vs deep networks has been in discussion for a long time in machine learning \cite{LarochelleH2007,Bengio+chapter2007} with pointers to the circuit complexity theory literature showing that shallow circuits can require exponentially more components than deeper circuits. The authors of residual networks tried to make them as thin as possible in favor of increasing their depth and having less parameters, and even introduced a <<bottleneck>> block which makes ResNet blocks even thinner.
\begin{figure}
  \centering
  \subfigure[basic]{\label{fig:blocks_basic}\includegraphics[scale=0.4]{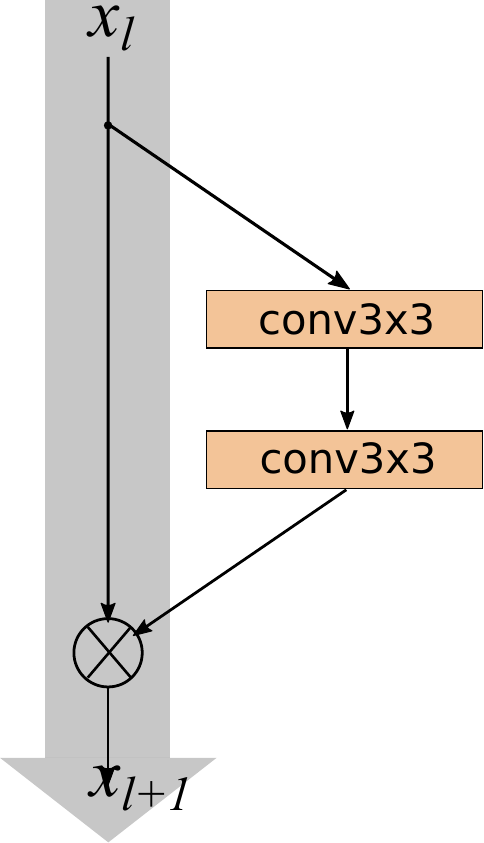}}
  \subfigure[bottleneck]{\label{fig:blocks_bottleneck}\includegraphics[scale=0.4]{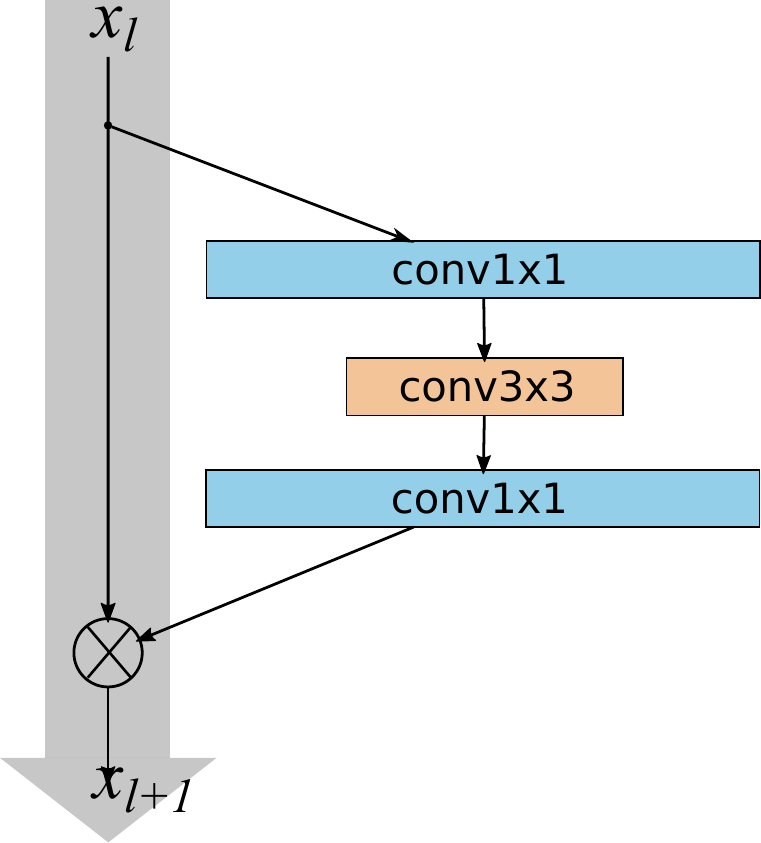}}
  \subfigure[basic-wide]{\label{fig:blocks_wide}\includegraphics[scale=0.4]{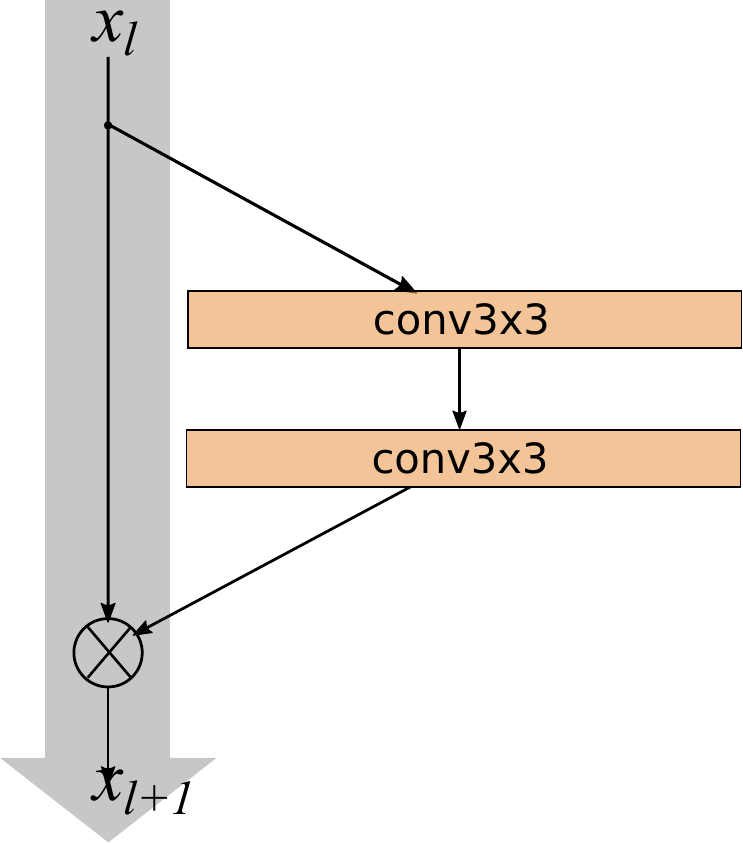}}
  \subfigure[wide-dropout]{\label{fig:blocks_dropout}\includegraphics[scale=0.4]{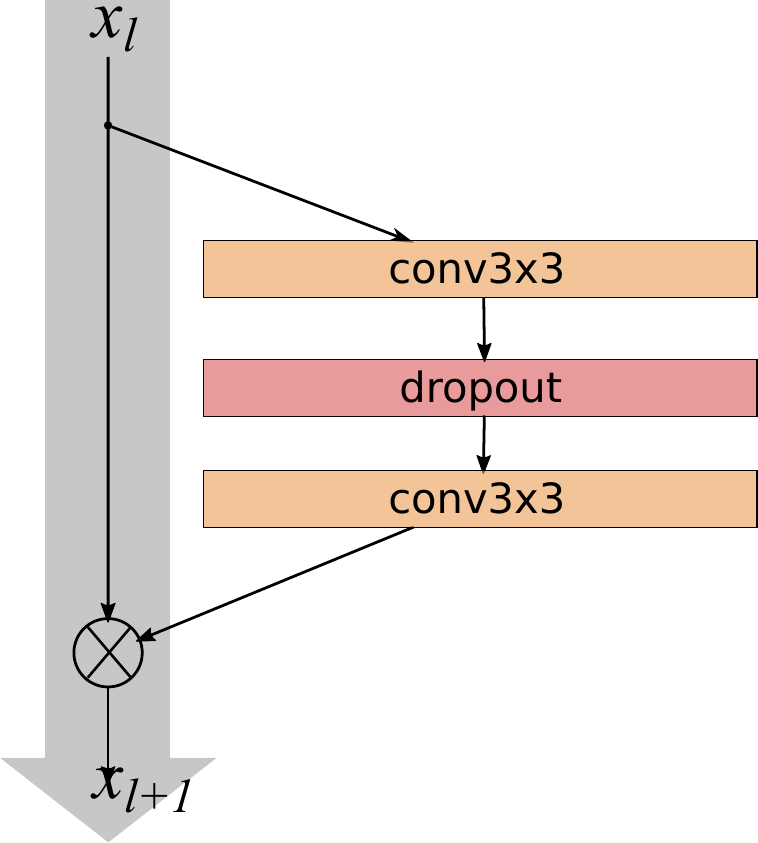}}
  \caption{Various residual blocks used in the paper. Batch normalization and ReLU precede each convolution (omitted for clarity)}
  \vspace{-0.4cm}
  \label{fig:blocks}
\end{figure}

 We note, however, that the residual block with identity mapping that allows to train very deep networks is at the same time a weakness of residual networks. As gradient flows through the network there is nothing to force it to go through residual block weights and it can avoid learning anything during training, so it is possible that there is either only a few blocks that learn useful representations, or many blocks share very little information with small contribution to the final goal. This problem was formulated as diminishing feature reuse in \cite{highway}.
The authors of \cite{stochastic_depth} tried to address this problem with the idea of randomly disabling residual blocks during training. This method can be viewed as a special case of dropout \cite{srivastava14a}, where each residual block has an identity scalar weight on which dropout is applied. The effectiveness of this approach proves the hypothesis above.

Motivated by the above observation, our work builds on top of \cite{basicblock2} and tries to answer the question of how wide deep residual networks should be and address the problem of training. In this context, we  show  that the  widening of  ResNet blocks (if done properly) provides a much more effective way of improving performance of residual networks compared to increasing their depth.
 In particular, we   present wider deep residual networks that significantly improve over \cite{basicblock2},  having \textit{50 times} less layers and being more than 2 times faster.
 We call the resulting network architectures \emph{wide residual networks}. For instance, our wide 16-layer deep network has the same accuracy as a 1000-layer thin deep network and a comparable number of parameters, although being several times faster to train. This type of experiments thus seem to indicate that the main power of deep residual networks is in residual blocks, and that the effect of depth is supplementary.
 We  note that one can train even better wide residual networks  that have twice as many parameters (and more), which suggests that to further improve performance by increasing depth of thin networks   one needs to add thousands of layers in this case.

\textbf{Use of dropout in ResNet blocks.}
Dropout was first introduced in \cite{srivastava14a} and then was adopted by many successful architectures as \cite{AlexNet,Simonyan15} etc. It was mostly applied on top layers that had a large number of parameters to prevent feature coadaptation and overfitting. It was then mainly substituted by batch normalization  \cite{icml2015_ioffe15} which was introduced as a technique to reduce internal covariate shift in neural network activations by normalizing them to have specific distribution. It also works as a regularizer and the authors experimentally showed that a network with batch normalization achieves better accuracy than a network with dropout. In our case, as widening of residual blocks results in an increase of the number of parameters, we studied the effect of dropout to regularize training and prevent overfitting. Previously, dropout in residual networks was studied in \cite{basicblock2} with dropout being inserted in the identity part of the block, and the authors showed negative effects of that. Instead, we argue here that dropout should be inserted between convolutional layers.
Experimental results on wide residual networks show that this leads to consistent  gains, yielding even new state-of-the-art results (\eg, 16-layer-deep wide residual network with dropout  achieves 1.64\% error on SVHN).

In summary, the contributions of this work are as follows:
\begin{itemize}
    \setlength\itemsep{0.1em}
  \item{We present a detailed  experimental study of residual network architectures that thoroughly examines several important aspects of ResNet block structure.}
  \item{We propose a novel \textit{widened} architecture for ResNet blocks that allows for residual networks with significantly improved  performance.}
  \item{We propose a new way of utilizing  dropout within deep residual networks so as to properly regularize them and prevent overfitting during training. }
  \item{Last, we show that our proposed ResNet architectures achieve state-of-the-art results on several datasets dramatically improving accuracy and speed of residual networks.}
\end{itemize}

\section{Wide residual networks}

Residual block with identity mapping can be represented by the following formula:

\begin{equation}
\ve{x}_{l+1} = \ve{x}_{l} + \mathcal{F}(\ve{x}_{l}, \mathcal{W}_l)
\end{equation}
where $x_{l+1}$ and $x_l$ are input and output of the $l$-th unit in the network, $\mathcal{F}$ is a residual function and $\mathcal{W}_l$ are parameters of the block. Residual network consists of sequentially stacked residual blocks.

In \cite{basicblock2} residual networks consisted of two type of blocks:

\begin{itemize}
  \item{\textit{basic} - with two consecutive $3\times3$ convolutions with batch normalization and ReLU preceding convolution: \texttt{conv$3\times3$-conv$3\times3$} Fig.\ref{fig:blocks_basic}}
  \item{\textit{bottleneck} - with one $3\times3$ convolution surrounded by dimensionality reducing and expanding $1\times1$ convolution layers: \texttt{conv$1\times1$-conv$3\times3$-conv$1\times1$} Fig.\ref{fig:blocks_bottleneck}}
\end{itemize}

Compared to the original architecture \cite{he2015deep} in \cite{basicblock2} the order of batch normalization, activation and convolution in residual block was changed from \texttt{conv}-\texttt{BN}-\texttt{ReLU} to \texttt{BN}-\texttt{ReLU}-\texttt{conv}. As the latter was shown to train faster and achieve better results we don't consider the original version.
Furthermore,
so-called  <<bottleneck>> blocks were initially used to make blocks less computationally expensive to increase the number of layers. As we want to study the effect of widening and <<bottleneck>> is used to make networks thinner we don't consider it too,  focusing instead on  <<basic>> residual architecture.

There are essentially three simple ways to increase representational power of residual blocks:
\vspace{-0.1cm}
\begin{itemize}
    \setlength\itemsep{-0.2em}
  \item{to add more convolutional layers per block}
  \item{to widen the convolutional layers by adding more feature planes}
  \item{to increase filter sizes in convolutional layers}
\end{itemize}

As small filters were shown to be very effective in several works including \cite{Simonyan15,inception-v4} we do not  consider using filters larger than $3\times3$.
Let us also introduce two factors, deepening factor $l$ and widening factor $k$, where $l$ is the number of convolutions in a block and $k$ multiplies the number of features in convolutional layers, thus the baseline <<basic>> block corresponds to  $l = 2$, $k = 1$. Figures \ref{fig:blocks_basic} and \ref{fig:blocks_wide} show schematic examples of <<basic>> and <<basic-wide>> blocks respectively.

\newcommand{\blocka}[2]{
  \(\left[
      \begin{array}{c}
        \text{3$\times$3, #1}\\[-.1em]
        \text{3$\times$3, #1}
      \end{array}
    \right]\)$\times$#2
}
\newcommand{\blockb}[2]{
  \(\left[
      \begin{array}{c}
        \text{3$\times$3, #1}\\[-.1em]
        \text{1$\times$1, #1}\\[-.1em]
        \text{3$\times$3, #1}
      \end{array}
    \right]\)$\times$#2
}
\newcommand{\convsize}[1]{#1$\times$#1}
\newcommand{\convname}[1]{#1}
\def\cellheight{0.34cm}
\begin{table}
  \centering
  \begin{tabular}{c|c|c}
    \hline
    group name & output size & block type = $B(3,3)$ \\
    \Xhline{2\arrayrulewidth}
    conv1 & $32\times32$ & [3$\times$3, 16] \\
    \convname{conv2} & \convsize{32} & \blocka{16$\times$k}{N} \\[\cellheight]
    \convname{conv3} & \convsize{16} & \blocka{32$\times$k}{N} \\[\cellheight]
    \convname{conv4} & \convsize{8} & \blocka{64$\times$k}{N} \\
    avg-pool & $1\times1$ & [$8\times8$] \\
    \hline
  \end{tabular}
  \vspace{0.2cm}
  \caption{Structure of wide residual networks. Network width is determined by factor $k$. Original architecture \cite{basicblock2} is equivalent to $k=1$. Groups of convolutions are shown in brackets where $N$ is a number of blocks in group, downsampling performed by the first layers in groups \texttt{conv3} and \texttt{conv4}. Final classification layer is omitted for clearance. In the particular example shown, the network uses  a ResNet block of type  $B(3,3)$. }
  \label{table:arch}
\end{table}

The general structure of our residual networks is illustrated in  table \ref{table:arch}: it consists of an initial convolutional layer \texttt{conv1} that is followed by 3 groups (each of size $N$) of residual blocks \texttt{conv2}, \texttt{conv3} and \texttt{conv4},  followed by average pooling and final classification layer. The size of \texttt{conv1} is fixed in all of our experiments, while the introduced widening factor $k$ scales the width of the residual blocks in the three groups \texttt{conv2-4} (\eg, the original <<basic>> architecture is equivalent to $k = 1$).
We want to study the effect of representational power of residual block and, to that end, we perform and test several modifications to the <<basic>> architecture, which are detailed in the following subsections.

\subsection{Type of convolutions in residual block}
Let $B(M)$ denote residual block structure, where $M$ is a list with the kernel sizes of the convolutional layers in a block. For example, $B(3,1)$ denotes a residual block with $3\times3$ and $1\times1$ convolutional layers (we always assume square spatial kernels). Note that, as  we do not consider <<bottleneck>> blocks as explained earlier, the number of feature planes is always kept the same across the block. We would like to answer the question of how important each  of the $3\times3$ convolutional layers of the <<basic>> residual architecture is and if they can be substituted by  a less computationally expensive $1\times1$ layer or even a combination of $1\times1$ and $3\times3$ convolutional layers,
\eg, $B(1,3)$ or $B(1,3)$. This can increase or decrease the representational power of the block. We thus experiment with the following combinations (note that the last combination, i.e., $B(3,1,1)$ is similar to effective Network-in-Network \cite{nin} architecture):

\vspace{-0.1cm}
\newcommand{\conv}[1]{C#1}
\begin{enumerate}
    \setlength\itemsep{-0.2em}
  \item{$B(3,3)$ - original <<basic>> block}
  \item{$B(3,1,3)$ - with one extra $1\times1$ layer}
  \item{$B(1,3,1)$ - with the same dimensionality of all convolutions, <<straightened>> bottleneck}
  \item{$B(1,3)$ - the network has alternating $1\times1$ - $3\times3$ convolutions everywhere}
  \item{$B(3,1)$ - similar idea to the previous block}
  \item{$B(3,1,1)$ - Network-in-Network style block}
\end{enumerate}

\subsection{Number of convolutional layers per residual block}
We also experiment with the block deepening factor $l$ to see how it affects performance.
The comparison has to be done among  networks with the same number of parameters, so in this case we need to build  networks with different $l$ and $d$ (where $d$ denotes the total number of blocks) while ensuring that  network complexity is kept roughly constant. This means, for instance, that $d$ should decrease whenever  $l$ increases.

\subsection{Width of residual blocks} In addition to the above modifications, we experiment with the widening factor $k$ of a block.
While the number of parameters increases linearly with $l$ (the deepening factor) and $d$ (the number of ResNet blocks), number of parameters and computational complexity are quadratic in $k$. However, it is more computationally effective to widen the layers than have thousands of small kernels as GPU is much more efficient in parallel computations on large tensors, so we are interested in an optimal $d$ to $k$ ratio.

One argument for wider residual networks would be that almost all architectures before residual networks, including the most successful Inception \cite{GoogLeNet} and VGG \cite{Simonyan15}, were much wider compared to \cite{basicblock2}. For example, residual networks WRN-22-8 and WRN-16-10 (see next paragraph for explanation of this notation) are very similar in width, depth and number of parameters to VGG architectures.

We further refer to original residual networks with $k=1$ as <<thin>> and to networks with $k>1$ as <<wide>>. In the rest of the paper we use the following notation: WRN-$n$-$k$  denotes a residual network that has a total number of convolutional layers $n$ and a widening factor $k$ (for example, network with 40 layers and $k=2$ times wider than original would be denoted as WRN-40-2). Also, when applicable we append block type, \eg WRN-40-2-$B(3,3)$.

\subsection{Dropout in residual blocks} As widening increases the number of parameters we would like to study ways of regularization. Residual networks already have batch normalization that provides a regularization effect, however it requires heavy data augmentation, which we would like to avoid, and it's not always possible. We add a dropout layer into each residual block between convolutions as shown in fig. \ref{fig:blocks_dropout} and after ReLU to perturb batch normalization in the next residual block and prevent it from overfitting. In very deep residual networks that should help deal with diminishing feature reuse problem enforcing learning in different residual blocks.

\vspace{-0.2cm}
\section{Experimental results}

For experiments we chose well-known CIFAR-10, CIFAR-100, SVHN and ImageNet image classification datasets. CIFAR-10 and CIFAR-100 datasets \cite{cifar10} consist of $32\times32$ color images drawn from 10 and 100 classes split into 50,000 train and 10,000 test images. For data augmentation we do horizontal flips and take random crops from image padded by 4 pixels on each side, filling missing pixels with reflections of original image. We don't use heavy data augmentation as proposed in \cite{fmp}. SVHN is a dataset of Google's Street View House Numbers images and contains about 600,000 digit images, coming from a significantly harder real world problem. For experiments on SVHN we don't do any image preprocessing, except dividing images by 255 to provide them in [0,1] range as input. All of our experiments except ImageNet are based on \cite{basicblock2} architecture with pre-activation residual blocks and we use it as baseline. For ImageNet, we find that using pre-activation in networks with less than 100 layers does not make any significant difference and so we decide to use the original ResNet architecture in this case.
 Unless mentioned otherwise, for CIFAR we follow the  image preprocessing of \cite{maxout} with ZCA whitening. However, for some CIFAR  experiments  we   instead   use simple mean/std normalization such that we can directly compare with  \cite{basicblock2} and other ResNet related works that make use of this type of preprocessing.

In the following we describe our findings w.r.t. the different  ResNet block architectures and also analyze the performance of our proposed wide residual networks.
We note that for all experiments related to <<type of convolutions in a block>> and <<number of convolutions per block>> we use  $k=2$ and reduced depth compared to \cite{basicblock2} in order to speed up training.\begin{table}
  \begin{tabular}{cc}
\begin{minipage}[c]{8cm}
  \centering
  \begin{tabular}{l|c|c|c|c}
    \hline
    block type & depth & \# params & time,s & CIFAR-10 \\
    \Xhline{2\arrayrulewidth}
$B(1,3,1)$ & 40 & 1.4M & 85.8 & 6.06  \\
$B(3,1)$ & 40 & 1.2M & 67.5 & 5.78  \\
$B(1,3)$ & 40 & 1.3M & 72.2 & 6.42  \\
$B(3,1,1)$ & 40 & 1.3M & 82.2 & 5.86  \\
$B(3,3)$ & 28 & 1.5M & 67.5 & 5.73  \\
$B(3,1,3)$ & 22 & 1.1M & 59.9 & 5.78  \\
    \hline
  \end{tabular}
  \caption{Test error (\%, median over 5 runs) on CIFAR-10 of residual networks with $k=2$ and different block types. Time column measures one training epoch.}
  \label{table:blocks}
\end{minipage}
\hspace{0.5cm}
\begin{minipage}[c]{4cm}
  \centering
  \begin{tabular}{c|c}
    \hline
    $l$ & CIFAR-10 \\
    \Xhline{2\arrayrulewidth}
    1 & 6.69 \\
    2 & 5.43 \\
    3 & 5.65 \\
    4 & 5.93 \\
    \hline
  \end{tabular}
  \caption{Test error (\%, median over 5 runs) on CIFAR-10 of WRN-40-2 (2.2M) with various $l$.}
  \label{table:blocks_l}
\end{minipage}
\end{tabular}
\end{table}

\subsubsection*{Type of convolutions in a block} We start by reporting results  using trained networks with different block types $B$ (reported results are on CIFAR-10). We used WRN-40-2 for blocks $B(1,3,1)$, $B(3,1)$, $B(1,3)$ and $B(3,1,1)$ as these blocks have only one $3\times3$ convolution. To keep the number of parameters comparable we trained other networks with less layers: WRN-28-2-$B(3,3)$ and WRN-22-2-$B(3,1,3)$. We provide the results including test accuracy in median over 5 runs and time per training epoch in the table \ref{table:blocks}. Block $B(3,3)$ turned out to be the best by a little margin, and $B(3,1)$ with $B(3,1,3)$ are very close to $B(3,3)$ in accuracy having less parameters and less layers. $B(3,1,3)$ is faster than others by a small margin.

Based on the above, blocks with comparable number of parameters turned out to give more or less the same results. Due to this fact, we hereafter restrict our attention to  only\ WRNs with $3\times3$ convolutions  so as to be also consistent with other methods.

\subsubsection*{Number of convolutions per block} We next proceed with the experiments related to varying the deepening factor $l$ (which represents the number of convolutional layers per block). We show indicative results in table~\ref{table:blocks_l}, where in this case we took WRN-40-2 with $3\times3$ convolutions and trained several networks with different deepening factor $l \in [1,2,3,4]$, same number of parameters (2.2$\times10^6$)  and same number of convolutional layers.

As can be noticed, $B(3,3)$ turned out to be the best, whereas $B(3,3,3)$ and $B(3,3,3,3)$ had the worst performance. We speculate that this is probably due to the increased difficulty in optimization as a result of the decreased number of residual connections in the last two cases. Furthermore, $B(3)$ turned out to be quite worse. The conclusion is that $B(3,3)$ is optimal in terms of number of convolutions per block. For this reason, in the remaining experiments we only consider wide residual networks with a block of type $B(3,3)$.

\subsubsection*{Width of residual blocks} As we try to increase widening parameter $k$ we have to decrease total number of layers. To find an optimal ratio we experimented with $k$ from 2 to 12 and depth from 16 to 40. The results are presented in table \ref{table:width}. As can be seen, all networks with 40, 22 and 16 layers see consistent gains when width is increased  by 1 to 12 times. On the other hand, when keeping  the same  fixed widening factor $k=8$ or $k=10$ and varying depth from 16 to 28 there is a consistent improvement, however when we further increase depth to 40 accuracy decreases (\eg, WRN-40-8 loses in accuracy to WRN-22-8).

We show additional results in table \ref{table:final} where we compare thin and wide residual networks. As can be observed, wide WRN-40-4 compares favorably to thin ResNet-1001 as it achieves better  accuracy on both CIFAR-10 and CIFAR-100. Yet, it is interesting that these networks have comparable number of parameters, {8.9$\times10^6$} and 10.2$\times10^6$, suggesting that depth does not add regularization effects compared to width at this level. As we show further in benchmarks, WRN-40-4 is 8 times faster to train, so evidently depth to width ratio in the original thin residual networks is far from optimal.

\begin{table}
  \centering
  \begin{tabular}{c|c|c|c|c}
\hline
depth & $k$ & \# params & CIFAR-10 & CIFAR-100 \\
\Xhline{2\arrayrulewidth}
40 & 1 & 0.6M & 6.85 & 30.89\\
40 & 2 & 2.2M & 5.33 & 26.04\\
40 & 4 & 8.9M & 4.97 & 22.89\\
40 & 8 & 35.7M & 4.66 & - \\
28 & 10 & 36.5M & \textbf{4.17} & 20.50\\
28 & 12 & 52.5M & 4.33 & \textbf{20.43}\\
22 & 8 & 17.2M & 4.38 & 21.22\\
22 & 10 & 26.8M & 4.44 & 20.75\\
16 & 8 & 11.0M & 4.81 & 22.07\\
16 & 10 & 17.1M & 4.56 & 21.59\\
\hline
  \end{tabular}
  \caption{Test error (\%) of various wide networks on CIFAR-10 and CIFAR-100 (ZCA preprocessing).}
  \label{table:width}
\end{table}

\begin{figure}
  \centering
  \includegraphics[scale=0.42]{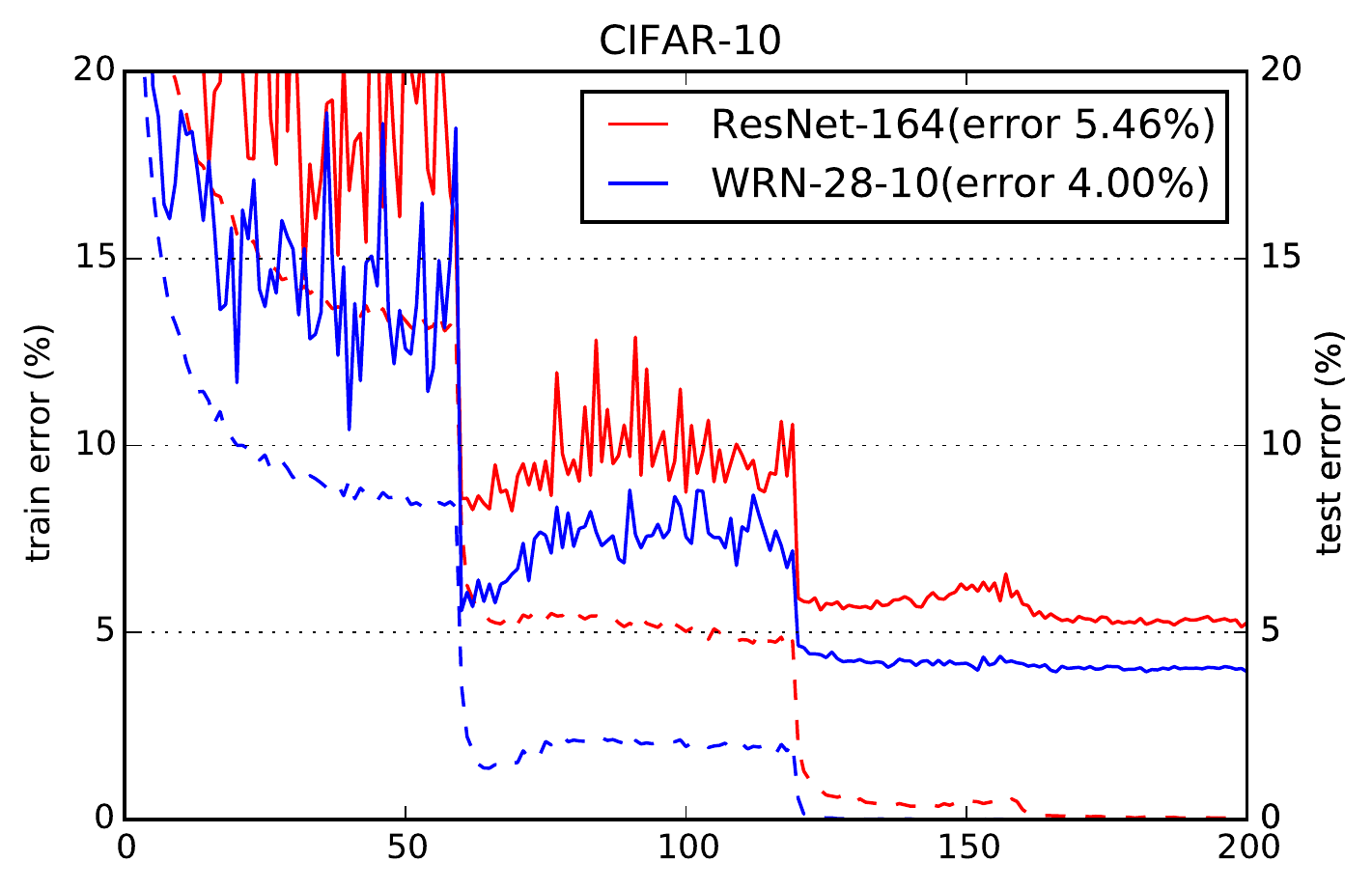}
  \includegraphics[scale=0.42]{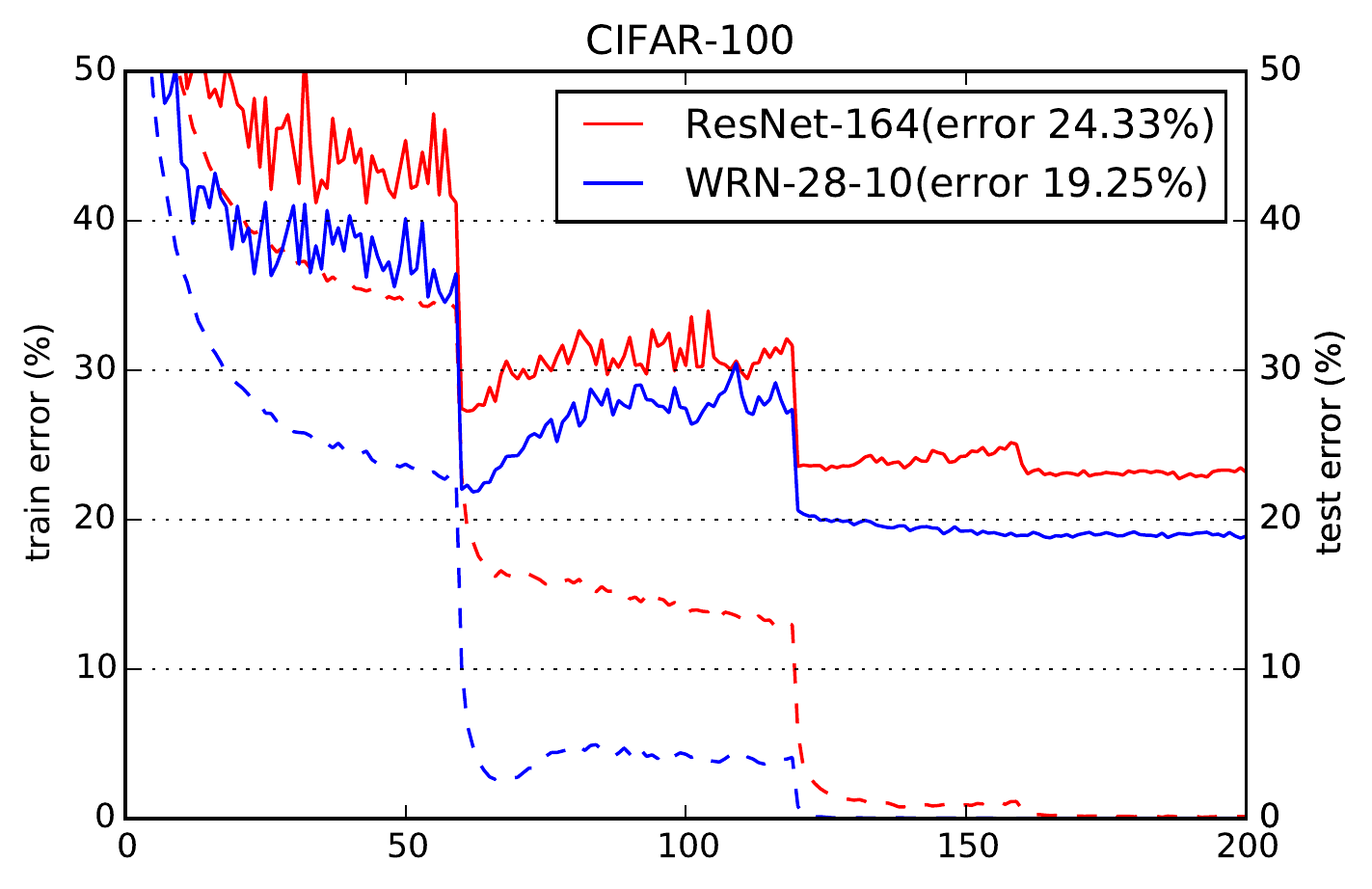} \caption{Training curves for thin and wide residual networks on CIFAR-10 and CIFAR-100. Solid lines denote test error (y-axis on the right), dashed lines denote training loss (y-axis on the left).}
  \label{fig:cifar}
\end{figure}

Also, wide WRN-28-10 outperforms thin ResNet-1001 by 0.92\% (with the same minibatch size during training) on CIFAR-10 and 3.46\% on CIFAR-100, having 36 times less layers (see table \ref{table:final}). We note that the result of 4.64\% with ResNet-1001 was obtained with batch size 64, whereas  we use a batch size 128 in all of our experiments (i.e., all other results reported in table \ref{table:final} are with batch size 128). Training curves for these networks are presented in Figure~\ref{fig:cifar}.

Despite previous arguments that depth gives regularization effects and width causes network to overfit, we successfully train networks with several times more parameters than ResNet-1001. For instance, wide WRN-28-10 (table \ref{table:final}) and wide WRN-40-10 (table \ref{table:overall}) have respectively $3.6$ and $5$ times more parameters than ResNet-1001 and both outperform it by a significant margin.

\begin{table}[ht]
  \centering
  \begin{tabular}{l|cccc}
    \hline
& depth-$k$ & \# params & CIFAR-10 & CIFAR-100 \\
\Xhline{2\arrayrulewidth}
NIN \cite{nin} & & & 8.81 & 35.67 \\
DSN \cite{dsn} & & & 8.22 & 34.57 \\
FitNet \cite{Romero-et-al-TR2014} & & & 8.39 & 35.04 \\
Highway \cite{highway} & & & 7.72 & 32.39 \\
ELU \cite{ELU} & & & 6.55 & 24.28 \\
\hline
\multirow{2}{*}{original-ResNet\cite{he2015deep}} & 110 & 1.7M & 6.43 & 25.16 \\
& 1202 & 10.2M & 7.93 & 27.82 \\
\hline
\multirow{2}{*}{stoc-depth\cite{stochastic_depth}} & 110 & 1.7M  & 5.23 & 24.58 \\
& 1202 & 10.2M & 4.91 & - \\
\hline
\multirow{3}{*}{pre-act-ResNet\cite{basicblock2}} & 110 & 1.7M & 6.37 & -\\
& 164 & 1.7M & 5.46 & 24.33 \\
& 1001 & 10.2M & 4.92(4.64) & 22.71 \\
\hline
\multirow{3}{*}{WRN (ours)} & 40-4 & 8.9M & 4.53 & 21.18 \\
& 16-8 & 11.0M & 4.27 & 20.43 \\
& 28-10 & 36.5M & \textbf{4.00} & \textbf{19.25} \\
\hline
  \end{tabular}
  \vspace{0.1cm}
  \caption{Test error of different methods on CIFAR-10 and CIFAR-100 with moderate data augmentation (flip/translation) and mean/std normalzation. We don't use dropout for these results. In the second column $k$ is a widening factor. Results for \cite{basicblock2} are shown with minibatch size 128 (as ours), and 64 in parenthesis. Our results were obtained by computing median over 5 runs.}
  \label{table:final}
  \vspace{-0.2cm}
\end{table}

In general,  we  observed that CIFAR mean/std preprocessing  allows training wider and deeper networks with better accuracy, and achieved 18.3\% on CIFAR-100 using WRN-40-10 with $56\times10^6$ parameters (table \ref{table:overall}), giving a total improvement of 4.4\% over ResNet-1001 and establishing a new state-of-the-art result on this dataset.

To summarize:
\begin{itemize}
  \item{widening consistently improves performance across residual networks of different depth;}
  \item{increasing both depth and width helps until the number of parameters becomes too high and stronger regularization is needed;}
  \item{there doesn't seem to be a regularization effect from very high depth in residual networks as wide networks with the same number of parameters as thin ones can learn same or better representations. Furthermore, wide networks can successfully learn with a 2 or more times larger number of parameters than thin ones, which would require doubling the depth of thin networks, making them infeasibly expensive to train.}
\end{itemize}

\subsubsection*{Dropout in  residual blocks} We trained networks with dropout inserted into residual block between convolutions on all datasets. We used cross-validation to determine dropout probability values, 0.3 on CIFAR and 0.4 on SVHN. Also, we didn't have to increase number of training epochs compared to baseline networks without dropout.

Dropout decreases test error on CIFAR-10 and CIFAR-100 by 0.11\% and 0.4\% correnspondingly (over median of 5 runs and mean/std preprocessing) with WRN-28-10, and gives improvements with other ResNets as well (table \ref{table:dropout}). To our knowledge, that was the first result to approach 20\% error on CIFAR-100, even outperforming methods with heavy data augmentation. There is only a slight drop in accuracy with WRN-16-4 on CIFAR-10 which we speculate is due to the relatively small number of parameters.

\begin{figure}
  \centering
  \vspace{0.1cm}
  \includegraphics[scale=0.42]{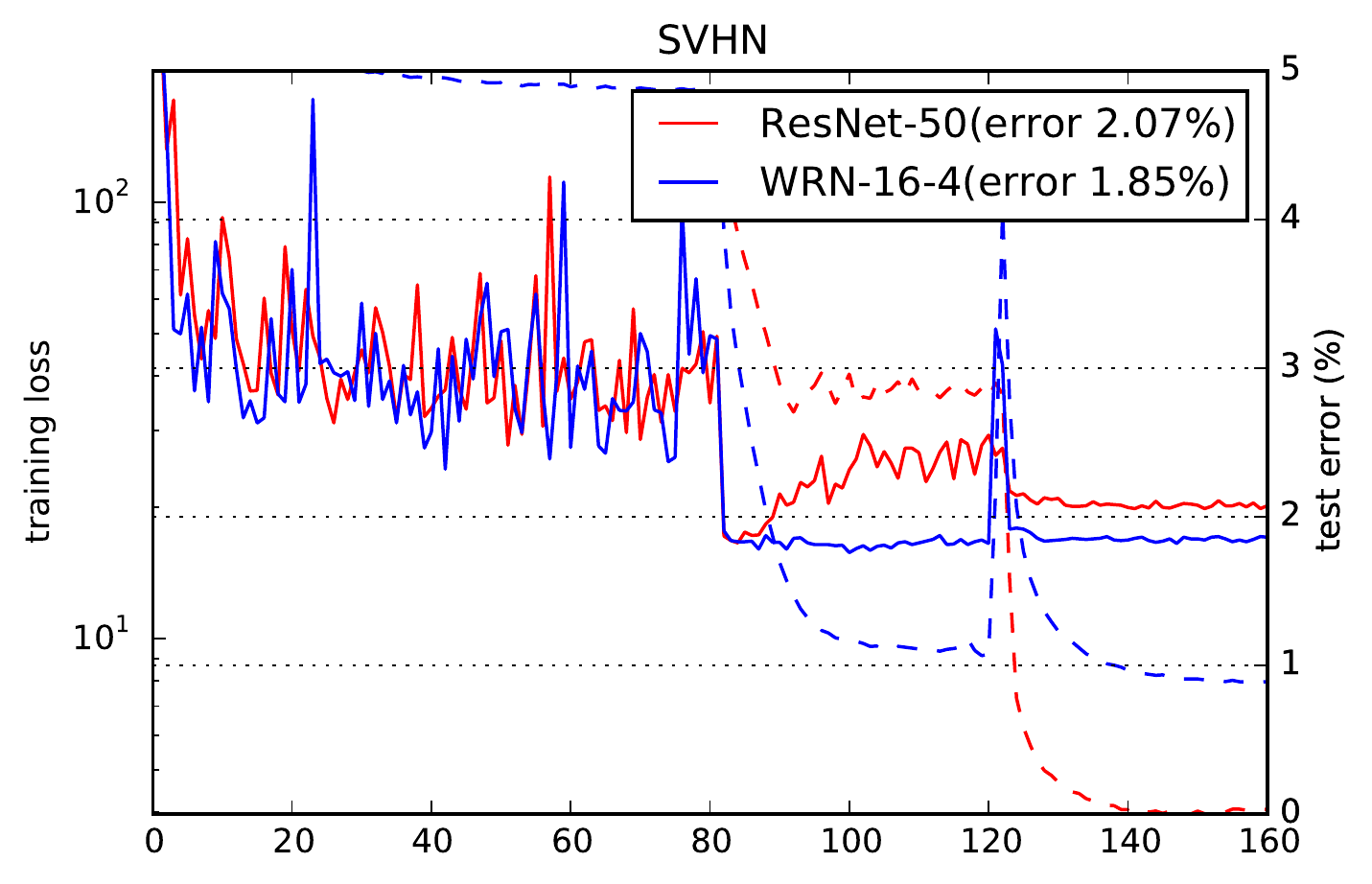}
  \includegraphics[scale=0.42]{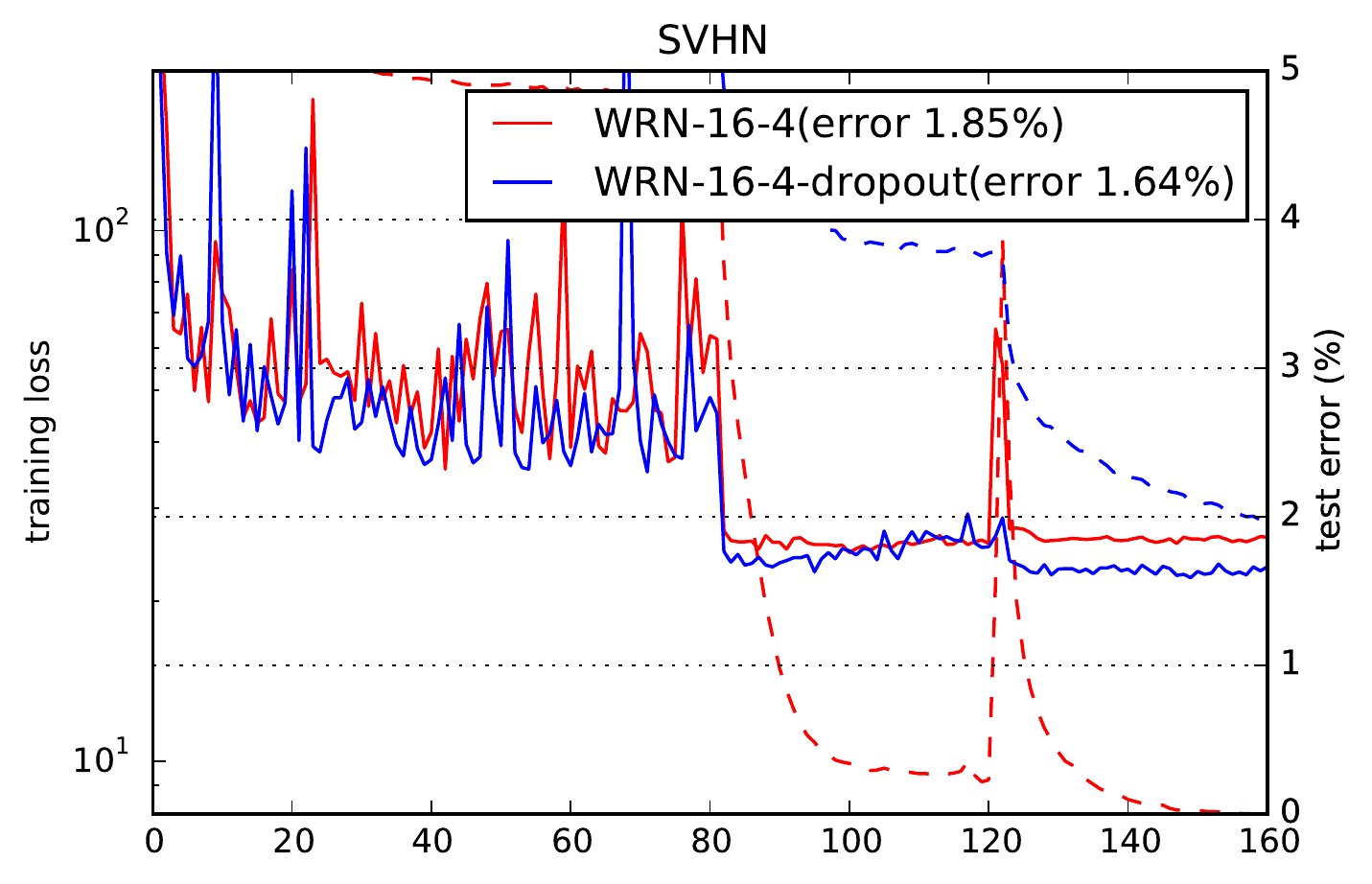}
  \caption{Training curves for SVHN. On the left: thin and wide networks, on the right: effect of dropout. Solid lines denote test error (y-axis on the right), dashed lines denote training loss (y-axis on the left).}
  \label{fig:dropout-svhn}
\end{figure}

We notice a disturbing effect in residual network training after the first learning rate drop when both loss and validation error suddenly start to go up and oscillate on high values until the next learning rate drop. We found out that it is caused by weight decay, however making it lower leads to a significant drop in accuracy. Interestingly, dropout partially removes this effect in most cases, see figures \ref{fig:cifar}, \ref{fig:dropout-svhn}.

The effect of dropout becomes more evident on SVHN. This is probably due to the fact that we don't do any data augmentation and batch normalization overfits, so dropout adds a regularization effect. Evidence for this can be found on training curves in figure \ref{fig:dropout-svhn} where the loss without dropout drops to very low values. The results are presented in  table \ref{table:dropout}. We observe significant improvements from using dropout on both thin and wide networks. Thin 50-layer deep network even outperforms thin 152-layer deep network with stochastic depth \cite{stochastic_depth}. We additionally trained WRN-16-8 with dropout on SVHN (table \ref{table:overall}), which achieves 1.54\% on SVHN - the best published result to our knowledge. Without dropout it  achieves 1.81\%.

Overall, despite the arguments of combining with batch normalization, dropout shows itself as an effective techique of regularization of thin and wide networks. It can be used to further improve results from widening, while also being complementary to it.

\newcommand{\chk}{{\centering\checkmark}}
\begin{table}
  \centering
  \begin{tabular}{c|c|c|c|c|c}
    \hline
     depth & $k$ & dropout & CIFAR-10 & CIFAR-100 & SVHN \\
    \Xhline{2\arrayrulewidth}
16 &     4 &      &     5.02 &   24.03 &         1.85\\
16 &     4 &     \chk & 5.24 &   23.91 &         1.64\\
28 &     10 &     &     4.00 &   19.25 & -\\
28 &     10 &    \chk & \textbf{3.89} &   \textbf{18.85} & -\\
52 &     1 &      &     6.43 &   29.89 &         2.08\\
52 &     1 &     \chk & 6.28 &   29.78 &         1.70\\
    \hline
  \end{tabular}
  \vspace{0.1cm}
  \caption{Effect of dropout in residual block. (mean/std preprocessing, CIFAR numbers are based on median of 5 runs)}
  \label{table:dropout}
\end{table}

\subsubsection*{ImageNet and COCO experiments}

For ImageNet we first experiment with non-bottleneck ResNet-18 and ResNet-34, trying to gradually increase their width from 1.0 to 3.0. The results are shown in table~\ref{table:imagenet_non_bottleneck}. Increasing width gradually increases accuracy of both networks, and networks with a comparable number of parameters achieve similar results, despite having different depth. Althouth these networks have a large number of parameters, they are outperfomed by bottleneck networks, which is probably either due to that  bottleneck architecture is simply better suited for ImageNet classification task, or due to that this more complex task needs a deeper network. To test this, we took the ResNet-50, and tried to make it wider by increasing inner $3\times3$ layer width. With widening factor of 2.0 the resulting WRN-50-2-bottleneck outperforms ResNet-152 having 3 times less layers, and being significantly faster. WRN-50-2-bottleneck is only slightly worse and almost $2\times$ faster than the best-performing pre-activation ResNet-200, althouth having slightly more parameters (table~\ref{table:imagenet_bottleneck}). In general, we find that, unlike CIFAR, ImageNet networks need more width at the same depth to achieve the same accuracy. It is however clear that it is unnecessary to have residual networks with more than 50 layers due to computational reasons.

We didn't try to train bigger bottleneck networks as 8-GPU machines are needed for that.

\begin{table}[ht]
  \centering\small
  \begin{tabular}{c|c|c|c|c|c}
    \hline
    \multicolumn{2}{c|}{width} & 1.0 & 1.5 & 2.0 & 3.0 \\ \hline
    \multirow{2}{*}{WRN-18} & top1,top5 & 30.4, 10.93 & 27.06, 9.0 & 25.58, 8.06 & 24.06, 7.33 \\
    & \#parameters & 11.7M & 25.9M & 45.6M & 101.8M \\
    \hline
    \multirow{2}{*}{WRN-34} & top1,top5 & 26.77, 8.67 & 24.5, 7.58 & 23.39, 7.00 &  \\
    & \#parameters & 21.8M & 48.6M & 86.0M \\
    \hline
  \end{tabular}
  \caption{ILSVRC-2012 validation error (single crop) of non-bottleneck ResNets for various widening factors. Networks with a comparable number of parameters achieve similar accuracy, despite having 2 times less layers.}
  \label{table:imagenet_non_bottleneck}
\end{table}

\begin{table}[ht]
  \centering\small
  \begin{tabular}{l|c|c|c|c}
    \hline
    Model & top-1 err, \% & top-5 err, \% & \#params & time/batch 16 \\ \hline
    ResNet-50 & 24.01 & 7.02 & 25.6M & 49 \\
    ResNet-101 & 22.44 & 6.21 & 44.5M & 82 \\
    ResNet-152 & 22.16 & 6.16 & 60.2M & 115 \\
    \bf{WRN-50-2-bottleneck} & 21.9 & 6.03 & 68.9M & 93 \\
    pre-ResNet-200 & 21.66 & 5.79 & 64.7M & 154 \\
    \hline
  \end{tabular}
  \caption{ILSVRC-2012 validation error (single crop) of bottleneck ResNets. Faster WRN-50-2-bottleneck outperforms ResNet-152 having 3 times less layers, and stands close to pre-ResNet-200.}
  \label{table:imagenet_bottleneck}
\end{table}

We also used WRN-34-2 to participate in COCO 2016 object detection challenge, using a combination of MultiPathNet \cite{Zagoruyko2016Multipath} and LocNet \cite{gidaris2016locnet}. Despite having only 34 layers, this model achieves state-of-the-art single model performance, outperforming even ResNet-152 and Inception-v4-based models. 

Finally, in table~\ref{table:overall} we summarize our best WRN results over various commonly used datasets.

\begin{table}[ht]
  \centering\small
  \begin{tabular}{l|l|c|c}
    \hline
    Dataset & model & dropout & test perf. \\
    \hline
    CIFAR-10  & WRN-40-10 & \chk & 3.8\% \\
    CIFAR-100 & WRN-40-10 & \chk & 18.3\% \\
    SVHN      & WRN-16-8 & \chk  & 1.54\% \\
    ImageNet (single crop) & WRN-50-2-bottleneck & & 21.9\% top-1, 5.79\% top-5 \\
    COCO test-std & WRN-34-2 & & 35.2 mAP \\
    \hline
  \end{tabular}
  \caption{Best WRN performance over various datasets, single run results. COCO model is based on WRN-34-2 (wider basicblock), uses VGG-16-based AttractioNet proposals, and has a LocNet-style localization part. To our knowledge, these are the best published  results for CIFAR-10, CIFAR-100, SVHN, and COCO (using non-ensemble models).}
  \label{table:overall}
\end{table}

\subsubsection*{Computational efficiency} 
\begin{figure}[h]
  \centering
  \includegraphics[scale=0.6]{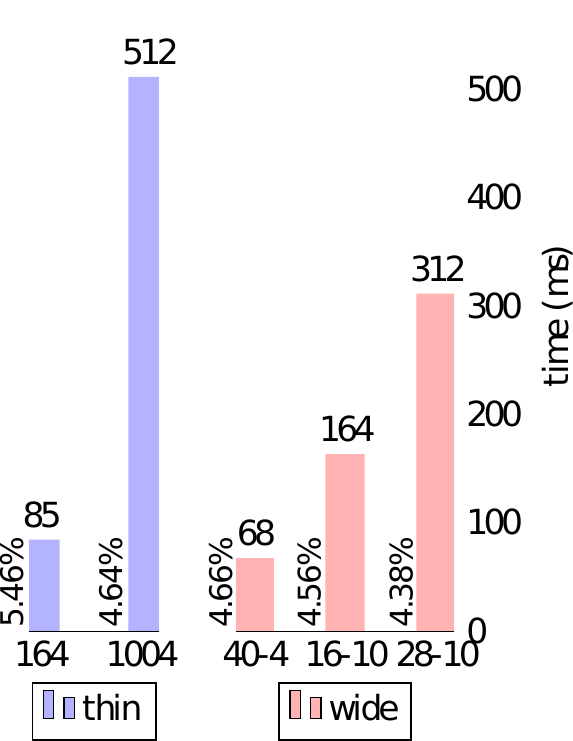}
  \caption{Time of forward+backward update per minibatch of size 32 for wide and thin networks(x-axis denotes network depth and widening factor). Numbers beside bars indicate test error on CIFAR-10, on top - time (ms). Test time is a proportional fraction of these benchmarks. Note, for instance, that wide WRN-40-4 is 8 times faster than thin ResNet-1001 while having approximately the same accuracy.
}
  \vspace{-0.2cm}
  \label{fig:benchmark}
\end{figure}

Thin and deep residual networks with small kernels are against the nature of GPU computations because of their sequential structure. Increasing width helps effectively balance computations in much more optimal way, so that wide networks are many times more efficient than thin ones as our benchmarks show. We use cudnn v5 and Titan X to measure forward+backward update times with minibatch size 32 for several networks, the results are in the figure \ref{fig:benchmark}. We show that our best CIFAR wide WRN-28-10 is 1.6 times faster than thin ResNet-1001. Furthermore, wide WRN-40-4, which has  approximately the same accuracy as ResNet-1001, is 8 times faster.

\subsubsection*{Implementation details} In all our experiments we use SGD with Nesterov momentum and cross-entropy loss. The initial learning rate is set to 0.1, weight decay to 0.0005, dampening to 0, momentum to 0.9 and minibatch size to 128. On CIFAR learning rate dropped by 0.2 at 60, 120 and 160 epochs and we train for total 200 epochs. On SVHN initial learning rate is set to 0.01 and we drop it at 80 and 120 epochs by 0.1, training for total 160 epochs. Our implementation is based on Torch \cite{torch}. We use \cite{optimize-net} to reduce memory footprints of all our networks. For ImageNet experiments we used \texttt{fb.resnet.torch} implementation \cite{fb.resnet.torch}. Our code and models are available at \url{https://github.com/szagoruyko/wide-residual-networks}.

\section{Conclusions}

We presented a study on the width of residual networks as well as   on the use of dropout in residual architectures. Based on this study, we  proposed a   wide residual network architecture that provides state-of-the-art results on several commonly used benchmark datasets (including CIFAR-10, CIFAR-100, SVHN and COCO), as well as significant improvements on ImageNet. We demonstrate that wide networks with only 16 layers can significantly outperform 1000-layer deep networks on CIFAR, as well as that 50-layer outperform 152-layer on ImageNet, thus showing that the main power of residual networks is in residual blocks, and not in extreme depth as claimed earlier. Also, wide residual networks are several times faster to train. We think that these intriguing findings will help further advances in research in deep neural networks.

\section{Acknowledgements}

We thank startup company VisionLabs and Eugenio Culurciello for giving us access to their clusters, without them ImageNet experiments wouldn't be possible. We also thank Adam Lerer and Sam Gross for helpful discussions. Work supported by EC project FP7-ICT-611145 ROBOSPECT.

\bibliography{bibliography}
\end{document}